\documentclass{article}

\usepackage[numbers,sort&compress]{natbib}

\usepackage[preprint,nonatbib]{neurips_2022}

\usepackage[utf8]{inputenc} 
\usepackage[T1]{fontenc}    
\usepackage{hyperref}       
\usepackage{url}            
\usepackage{booktabs}       
\usepackage{nicefrac}       
\usepackage{microtype}      
\usepackage{xcolor}         

\usepackage{graphicx}
\usepackage{caption}        
\usepackage{subcaption}
\usepackage{wrapfig}        
\usepackage{amsmath,amsfonts,amssymb} 
\usepackage{bbm}            

\title{Graph Contrastive Learning for Materials}

\author{
   Teddy Koker
   \qquad Keegan Quigley \qquad Will Spaeth \qquad Nathan C. Frey \qquad Lin Li \\
   MIT Lincoln Laboratory \\
   Lexington, MA 02421-6426 \\
 \texttt{\{thomas.koker,keegan.quigley,john.spaeth,nathan.frey,lin.li\}@ll.mit.edu} \\
}

\begin{document}

\maketitle

\begin{abstract}
    Recent work has shown the potential of graph neural networks to efficiently predict material properties, enabling
    high-throughput screening of materials. Training these models, however, often requires large quantities of labelled data, obtained via
    costly methods such as ab initio calculations or experimental evaluation. By leveraging a series of material-specific transformations, we introduce CrystalCLR, a framework for constrastive learning of representations with crystal graph neural networks. With the addition of a novel loss function, our framework is able to learn representations competitive with engineered fingerprinting methods. We also demonstrate that via model finetuning, contrastive pretraining can improve the performance of graph neural networks for prediction of material properties and significantly outperform traditional ML models that use engineered fingerprints. Lastly, we observe that CrystalCLR produces material representations that form clusters by compound class.
    
\end{abstract}


\section{Introduction}
The discovery of novel materials is a problem of considerable interest. Machine learning (ML) has demonstrated potential in accelerating material discovery with accurate prediction performance and lower computational cost \cite{Schleder2019}. Traditional ML models used in material discovery rely on expert knowledge to construct material fingerprints as inputs to ML models \cite{Ramprasad2017}. The development of fingerprint-based representations require manual design which is often limited to specific material families and properties and can be time-consuming to compute.

Recent advances in deep learning provide an alternative way of learning feature representations automatically from data. In particular, graph neural networks (GNNs) have been developed for modeling crystalline materials to predict material properties \cite{xie2018crystal,Chen2019,Louis2020,griggs2020unified,batzner20223,musaelian2022learning}. In addition, \citet{magar2021auglichem} have introduced several augmentations
specific to crystalline structures, including perturb-structure and supercell. They find that training GNNs on augmented crystal data results in improved performance in
prediction of several material properties. While GNNs are becoming increasingly popular, like
in other deep learning domains, adoption is limited by the availability of high quality
labelled data \cite{fung2021benchmarking}. Oftentimes, we only have access to very limited labelled data, such as materials' physical properties (e.g., melting point, thermal conductivity) that require laborious and time-consuming experimentation or computationally expensive {\it ab initio} calculations. One way to tackle this problem is to transfer the learned model trained on a related base task with large labelled data to the target task \cite{lee2021transfer}. However, this approach is limited by the availability of large labelled data for the base task. In the absence of labelled data, self-supervised methods provide a way to learn representations from samples alone.



Contrastive learning is one such method of self-supervised learning, which seeks to learn representations such that similar pairs remain close in embedding space, while others remain distant. In recent years, this has become a powerful method of both learning of
representations, as well as self-supervised pre-training for transfer learning, primarily in the image domain \cite{chen2020simple}. \citet{you2020graph} introduced a framework for applying
contrastive learning to graphs, using GNNs and broadly applicable transformations to biochemical
molecules, social network, and image super-pixel graphs. Similar work has shown that contrastive
pretraining can improve supervised performance in predicting the chemical properties of molecules \cite{wang2022molecular}. Lastly, \citet{khosla2020supervised} showed how supervised labels can be incorporated into the contrastive learning framework, allowing these labels to be incorporated into the representations. Despite the success of contrastive learning in a broad range of domains, very little work has been done to apply contrastive learning to crystalline materials.  

In this work, we introduce CrystalCLR, a framework for the constrastive representation learning of crystalline materials.
We highlight the importance of augmentation selection, and show that contrastive learning is
capable of learning representations competitive with fingerprinting methods. We also incorporate a novel loss function into the contrastive learning objective, which further improve representation quality. Furthermore, we show transfer learning with the pretrained contrastive model can outperform supervised learning for the prediction of material properties.

\section{Method}


Following \cite{chen2020simple,you2020graph}, we construct our constrastive learning framework with four major components: \textit{augmentation}, \textit{GNN encoder}, \textit{projection head} and \textit{contrastive loss}. 
Let $V$ and $E$ denote the atomic and bond attributes of a material's crystal structure, respectively. Each crystal graph $\mathcal{G}=(V,E)$ is first augmented into a similar pair $\hat{\mathcal{G}}_i$ and $\hat{\mathcal{G}}_j$ where  $\hat{\mathcal{G}}_i=(\hat{V}_i,\hat{E}_i)$ and the augmentations used to transform $\mathcal{G}$ into $\hat{\mathcal{G}}$ are specific to the domain of crystalline materials (Sec. \ref{augmentations}). 
A \textit{GNN encoder}, $f(\cdot)$, maps crystal graphs $\hat{\mathcal{G}}_i,\hat{\mathcal{G}}_j$ into 
representations $\boldsymbol{h}_i,\boldsymbol{h}_j$.
For our work we use the CGCNN architecture as the encoder, following the same graph representation of
crystal structure as inputs \cite{xie2018crystal}.
The \textit{projection head}, a two layer MLP $g(\cdot)$, projects representations $\boldsymbol{h}_i,\boldsymbol{h}_j$ into 128-dimensional space to create projections $\boldsymbol{z}_i,\boldsymbol{z}_j$. The addition of a non-linear projection prior to the loss function has shown to improve representation quality \cite{chen2020simple}.
The \textit{contrastive loss} function seeks to maximize the agreement between representations $\boldsymbol{z}_i,\boldsymbol{z}_j$ augmented from the same crystalline material while minimizing the agreement among the rest of the pairs augmented from different crystalline materials. We use the NT-Xent loss \cite{sohn2016improved}, written for pair $\boldsymbol{z}_i,\boldsymbol{z}_j$ in a batch of $N$ pairs as:


%
%

\begin{equation}\label{eq:1}
\mathcal{L}^\text{CLR}_{i} = -\log \frac
{\exp(\textrm{sim}(\boldsymbol{z}_i, \boldsymbol{z}_j) / \tau)}
{\sum^{2N}_{k=1}\mathbbm{1}_{[k \neq i]}\exp(\textrm{sim}(\boldsymbol{z}_i, \boldsymbol{z}_k)/ \tau)}
\end{equation}

where $\textrm{sim}(\boldsymbol{u},\boldsymbol{v})$ denotes cosine similarity 
($\boldsymbol{u}^\top \boldsymbol{v} / \lVert \boldsymbol{u} \rVert \lVert \boldsymbol{v} \rVert$). The final loss is the sum across all positive pairs in the batch.

\subsection{Crystal Augmentations}\label{augmentations}

\begin{figure}[!t]
\centering
    \begin{subfigure}{0.19\textwidth}
        \centering
        \includegraphics[width=\linewidth]{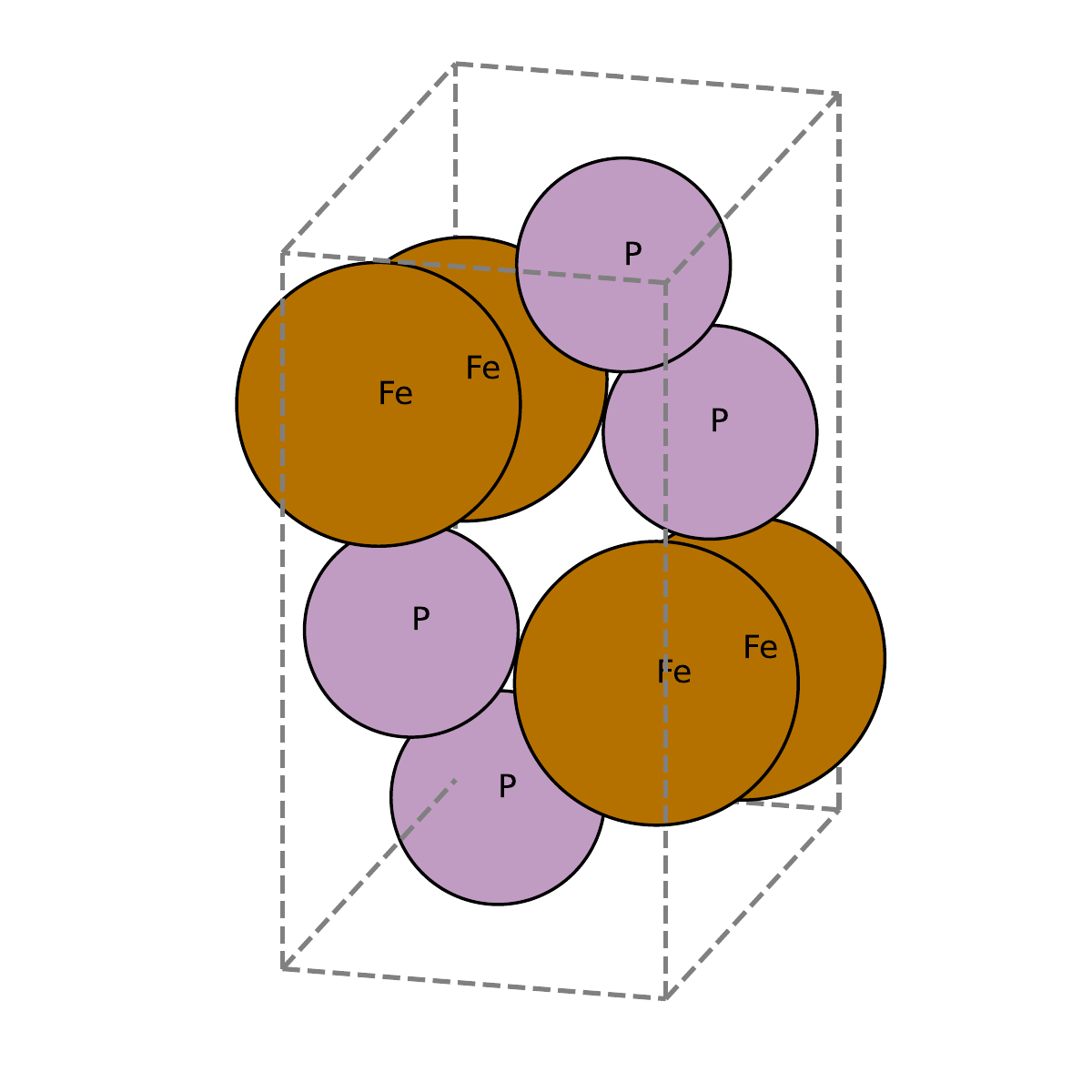}
        \caption{Original}
    \end{subfigure}
    \begin{subfigure}{0.19\textwidth}
        \centering
        \includegraphics[width=\linewidth]{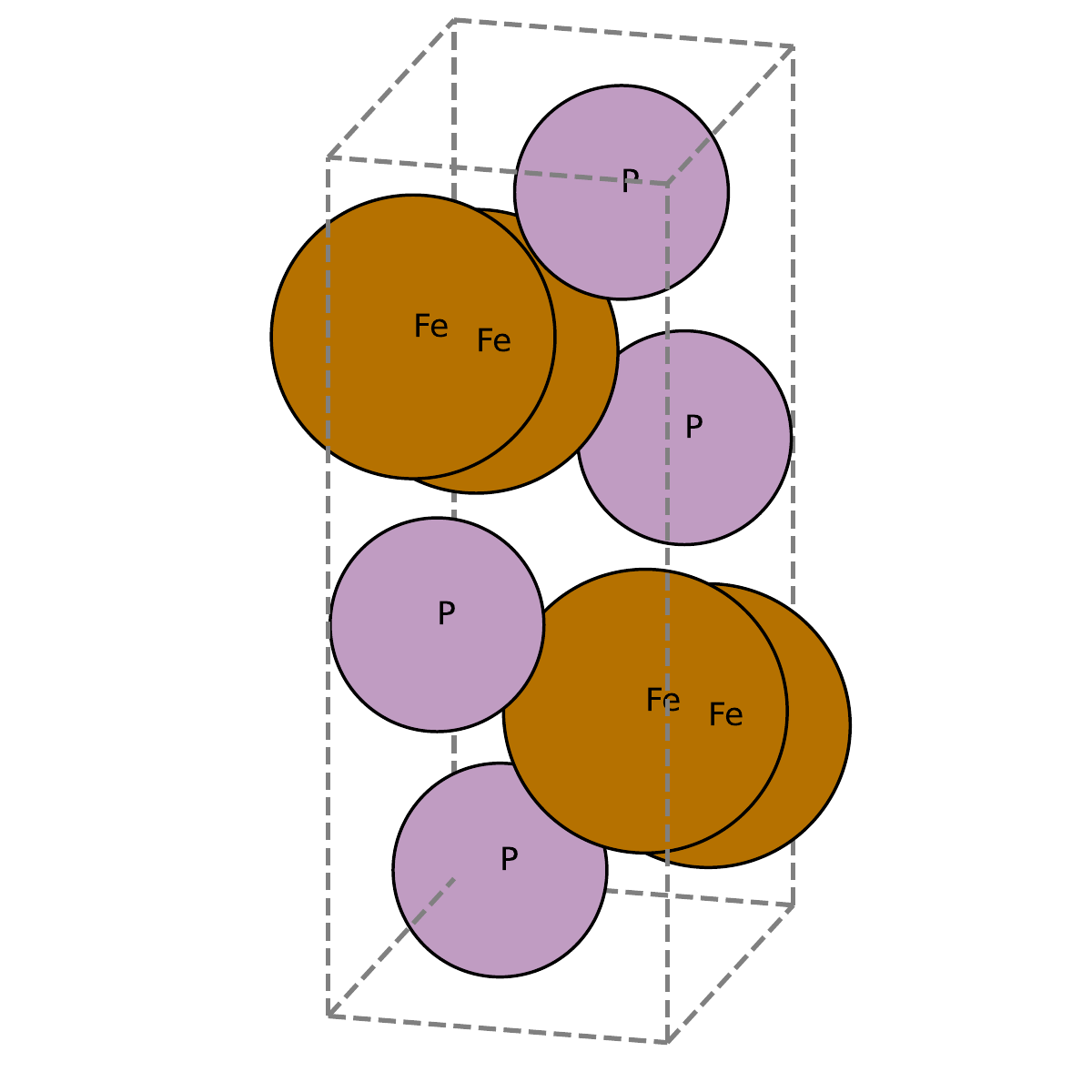}
        \caption{Strain}
    \end{subfigure}
    \begin{subfigure}{0.19\textwidth}
        \centering
        \includegraphics[width=\linewidth]{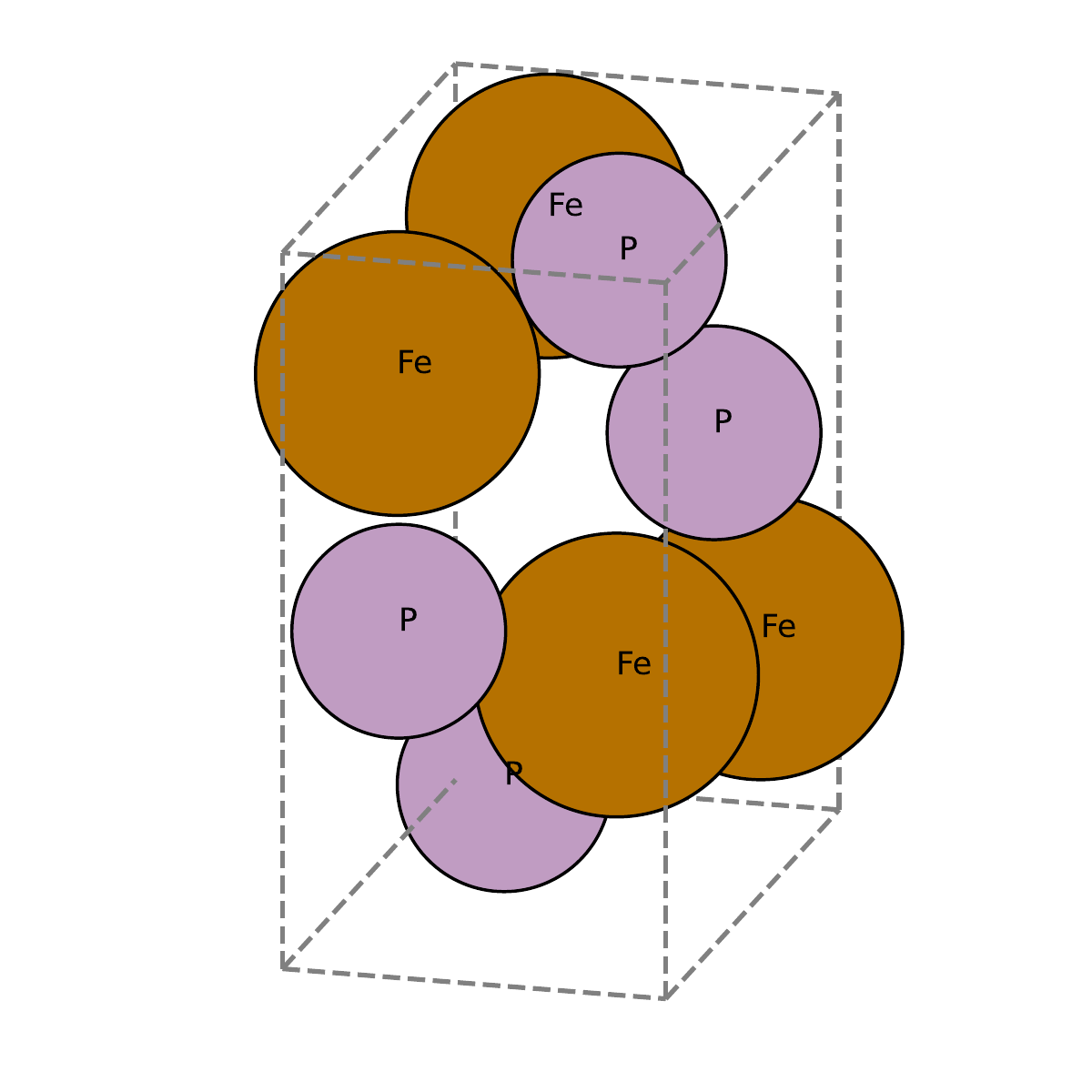}
        \caption{Perturb}
    \end{subfigure}
    \begin{subfigure}{0.19\textwidth}
        \centering
        \includegraphics[width=\linewidth]{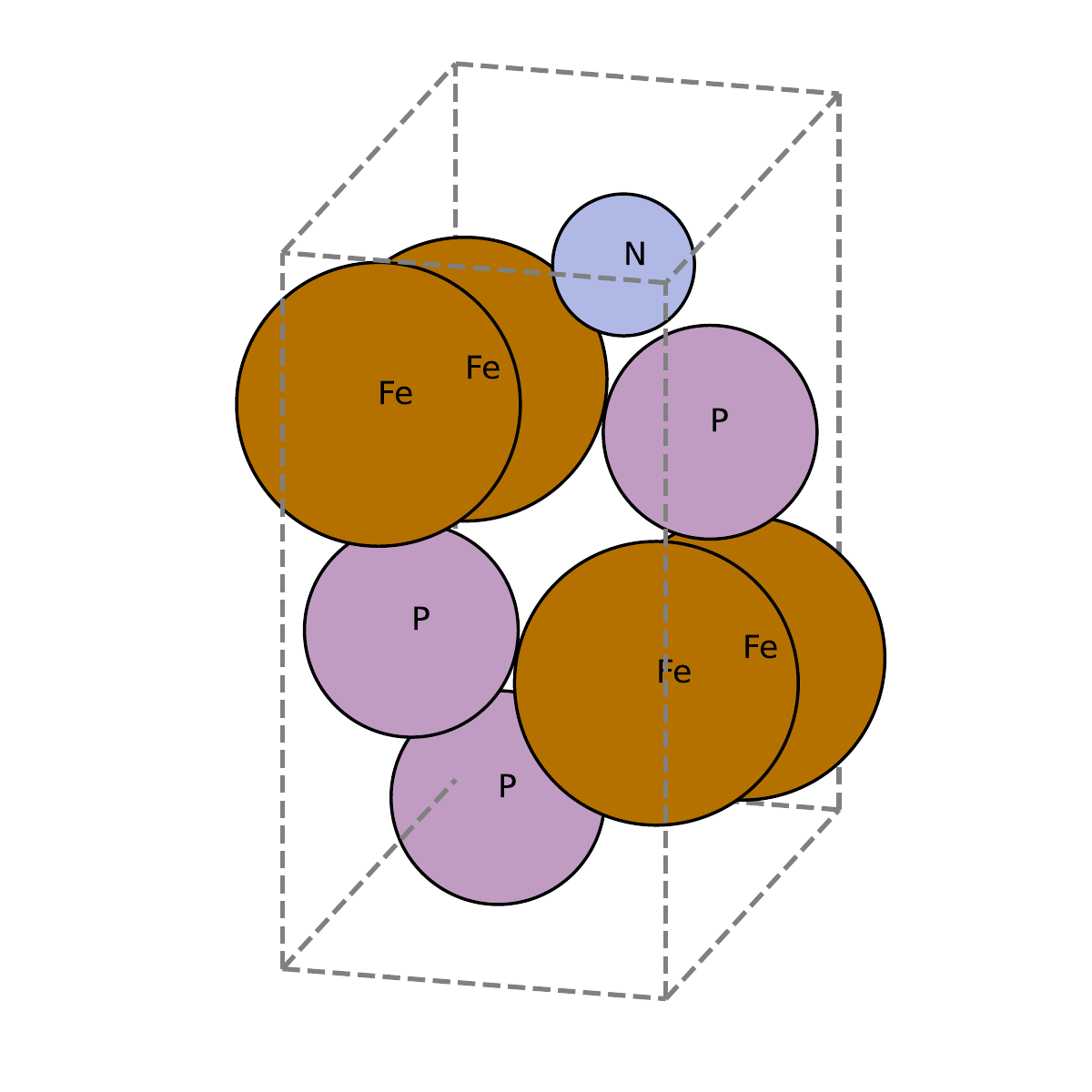}
        \caption{Column replace}
    \end{subfigure}
    \begin{subfigure}{0.19\textwidth}
        \centering
        \includegraphics[width=\linewidth]{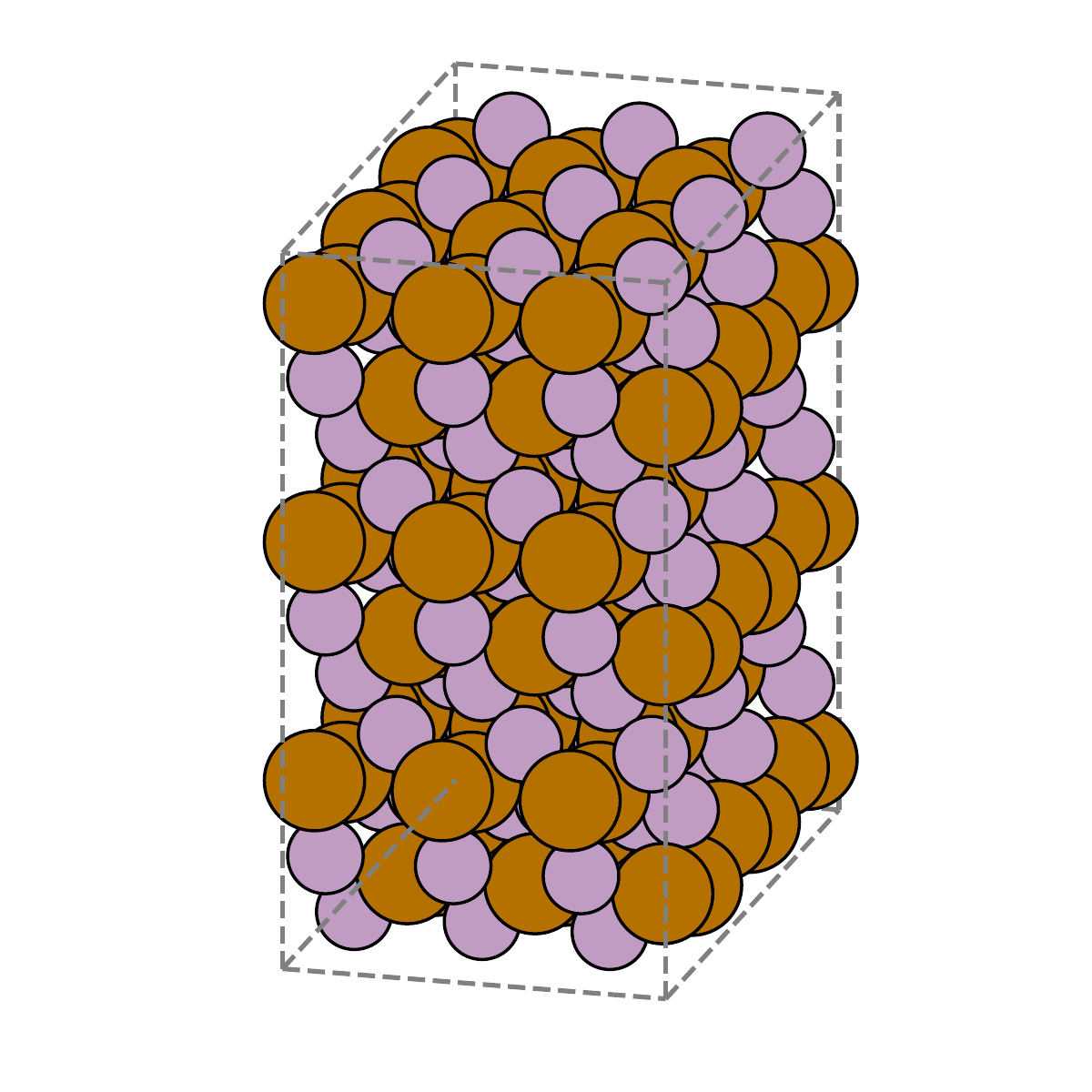}
        \caption{Supercell}
    \end{subfigure}
    \caption{Visualizations of the crystal transformations used. All transforms are stochastic by nature with the exception of supercell.}
    \label{fig:data_aug}
    \vspace{-0.5cm}
\end{figure}

We investigate four transformations specific to crystalline materials (Fig. \ref{fig:data_aug}). Each transformation is isoelectronic, preserving the valence electron configuration of the material: (1) \textit{perturb structure}: each site is perturbed by a distance uniformly sampled between 0 and 50\% of the minimum pairwise distance within the structure; (2) \textit{strain}: a random, anisotropic tensile strain deformation is applied to the structure. Each lattice vector is increased by a factor uniformly sampled from $[0,0.05]$; (3) \textit{column replace}: with probability $1$, a single, randomly selected site within the structure is replaced with an element from the same column in the periodic table as the original element; and (4) \textit{supercell}: a supercell of the original crystal is created, scaling each lattice vector by
a factor of $3$. The \textit{perturb structure} strength range is selected based on
values used in prior work \cite{magar2021auglichem}, and the \textit{strain}
strength range is selected based on physically reasonable values.
\subsection{Optimizing for composition similarity} 

Under the notion that materials with similar composition may tend to have similar properties, we hypothesize that explicitly guiding representations of similarly composed materials together may improve the representation quality.
We propose an additional loss term, composition similarity (CS), to promote representation similarity among
materials that
have one or more common elements. Each crystal is assigned a binary vector $a \in \{0, 1\}^{100}$, indicating
which of the possible 100 atoms (that exist in the dataset used) are contained in the crystal. The set
$P(i) = \{ p \in I : a_i \cdot a_p > 0, i \neq p \}$ is the set of indices that share one or more atom with the $i^{th}$
crystal. The loss can then be defined as:

\begin{equation}\label{eq:2}
\mathcal{L}^\text{CS}_{i} = -\log \Bigl\{\frac{1}{|P(i)|} \sum_{j \in P(i)} \frac
{\exp(\text{sim}(z_i, z_j) / \tau )}
{\sum^{2N}_{k=1}\mathbbm{1}_{[k \neq i]} \exp(\text{sim}(z_i, z_k)/ \tau )}
\Bigr\}
\end{equation}

Where $|P(i)|$ is the cardinality of the set. While prior work in \cite{khosla2020supervised} has shown that
placement of the summation over positives \textit{outside} the $\log$ is more theoretically optimal, we found the above formulation to produce better representations
empirically.
In the final model, \eqref{eq:2} is used in combination with
the original NT-Xent loss \eqref{eq:1} and weighted equally.

\subsection{Learning Task Description}
Our self-supervised training dataset consists of 90,160 crystal structures
downloaded from Materials Project in April, 2021 \cite{jain2013commentary}. 
For evaluation, we use additional melting temperature, thermal conductivity, and bulk moduli (K\_VRH) datasets of sizes
3,014, 5,540, and 13,121 respectively. The melting temperature data were experimental data scraped from MatWeb \cite{gao2013recent}. Thermal conductivity data were downloaded from AFLOW \cite{Curtarolo2012}; all values are at 300K. Bulk moduli data were downloaded from Materials Project. Each dataset is split 80/10/10 into train, test, and validation sets.

\section{Experimental Setup and Results}

\subsection{Augmentation Study}
\begin{wrapfigure}{r}{0.5\textwidth}
  \vspace{-2.3cm} 
  \centering
  \includegraphics[width=\linewidth]{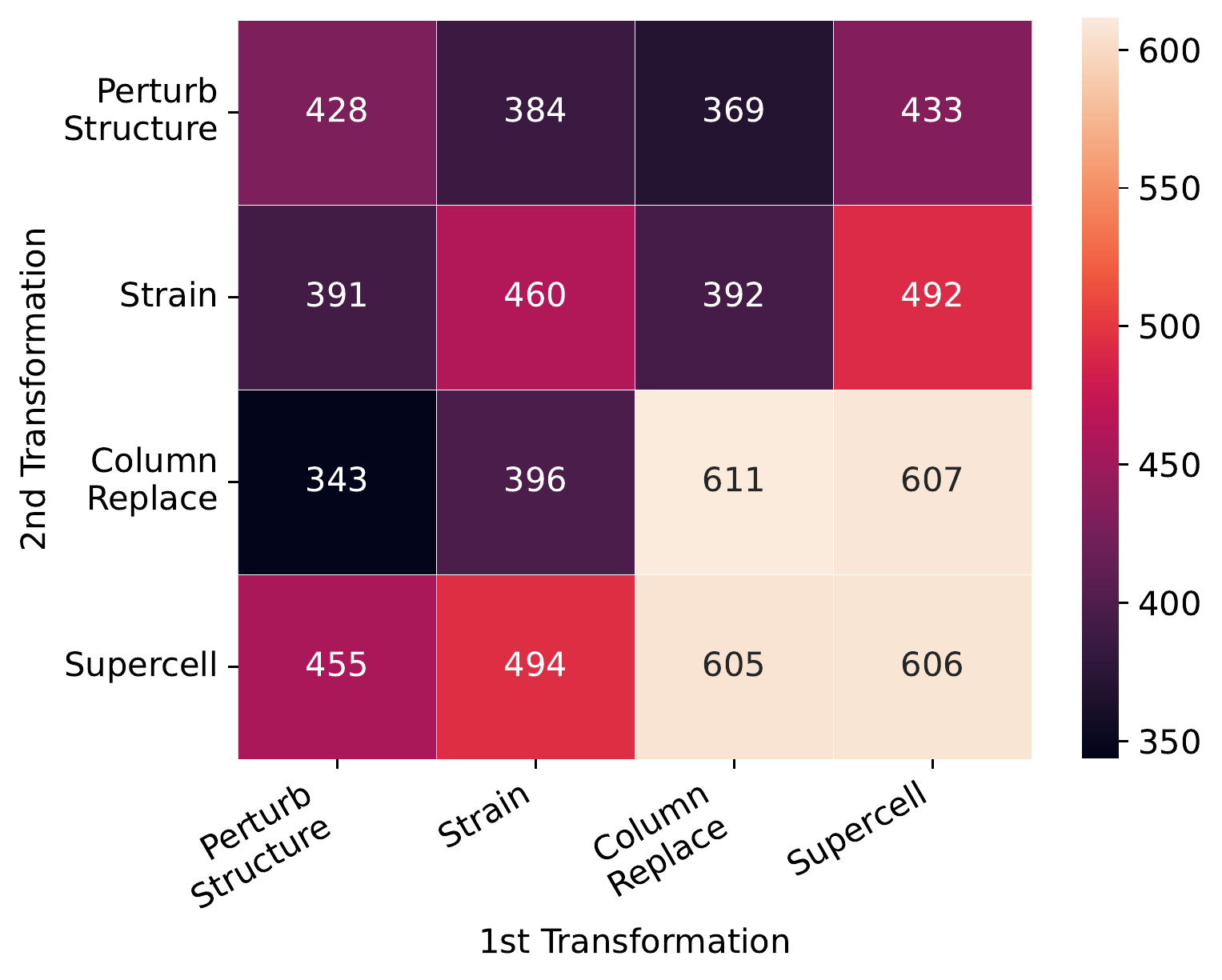}
  \vspace{-0.7cm}
  \caption{Linear regression evaluation of mean-absolute-error (MAE) of melting temperature. Single transformations are applied for elements along the diagonal, while other elements contain two transformations applied sequentially.}
  \vspace{-0.3cm} 
  \label{fig:aug_study}
\end{wrapfigure}

To evaluate the effects of individual and combinations of augmentations, we train models under all of the pairwise
combinations of augmentations. For this study, we apply augmentations to only one sample in each positive pair.
Each model is evaluated by performing a linear regression on the learned representations for each of the studied material properties. In our work, we use no augmentations for evaluation.
Similar to \cite{chen2020simple,you2020graph}, we observe that the composition of transformations
is crucial for representation quality. As shown in Figure \ref{fig:aug_study} for melting point prediction, the best
representations are obtained using the perturb-structure and column-replace transformations, and the addition of supercell often deteriorates performance. We observe a similar trend across all of the studied material properties.

\subsection{Model Evaluation}
We selected perturb-structure, strain, and column-replace as the final set of data augmentations; each augmentation has a 50\% chance of being applied to each training sample. We trained two self-supervised models; each model was trained for 5000 epochs, with a batch size of $512$. The first model (CrystalCLR) uses
$\mathcal{L}^\text{CLR}$ alone, while the second model (CrystalCLR+CS) uses a final loss of 
$\mathcal{L}^\text{CLR} + \mathcal{L}^\text{CS}$. We use the Adam optimizer \cite{kingma2014adam}
with a learning rate of 1e-4.

To quantitatively evaluate the learned representations, we trained random-forest regression models \cite{breiman2001random} to predict
material properties. We compare the predictive performance of the learned embeddings to several other material fingerprinting
methods, all obtainable without supervision: 
(1) Site stats fingerprint that aggregates statistics of the CrystalNNFingerprint \cite{zimmermann2020local} using structural order parameters, (2) Sine Coulomb
fingerprint which is a variant of the Coloumb matrix for periodic crystals \cite{faber2015crystal},
and (3) Element property which is a weighted mean, standard deviation, minimum and maximum
of element properties obtained from Magpie \cite{ward2016general}. See Table \ref{table:element_properties} for the list of properties used. We use the default hyperparameters for the
random forest from \texttt{scikit-learn} \cite{scikit-learn}.

Table \ref{table:results} shows the representations learned from contrastive training perform at or above traditional fingerprinting
methods, with the exception of element properties. Element properties contain additional information, including melting point, which explain its high performance. Furthermore, we observe that the addition of the composition loss yields an improvement in performance across all properties. 

%


\begin{figure}[h]
    \hspace{-1.1cm}
    \vspace{-0.1cm} 
    \centering
    \includegraphics[width=\linewidth]{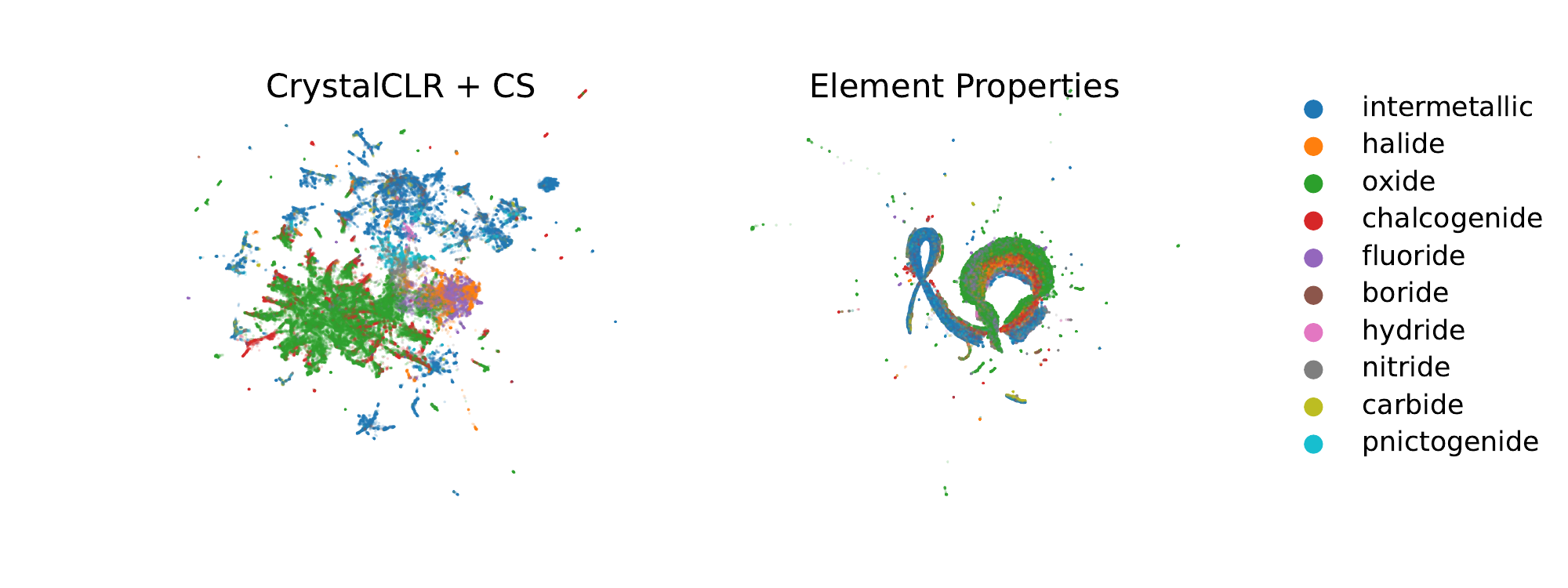}
    \vspace{-0.9cm}
    \caption{UMAP \cite{mcinnes2018umap} visualizations of CrystalCLR + CS and Element Properties embeddings, colored by compound classes.}
    \label{fig:embeddings}
    \vspace{0.2cm}
\end{figure}


We also evaluate the use of contrastive learning as a pretraining task for material property prediction. For each of aforementioned properties, we train CGCNN models; one randomly initialized, and one initialized with the CGCNN encoder weights from each of the two contrastive
training methods. For each model, we use the Adam optimizer \cite{kingma2014adam} to fine-tune the models, using the learning rate with the best validation performance. 

Table \ref{table:results} shows by transferring weights from the contrastive-pretrained encoder, we are able to obtain more accurate
predictions of material properties than a randomly initialized model for melting temperature and bulk moduli. The benefit of the
composition loss is more evident for melting temperature prediction than the thermal conductivity and bulk moduli properties.

Lastly, in Figure \ref{fig:embeddings} we visualize produced crystal embeddings using UMAP \cite{mcinnes2018umap}. CrystalCLR learns representations that form clusters by compound class, despite
training in a self-supervised manor.

\begin{table}
  \centering
  \begin{tabular}{llll}
    \toprule
        & Melting temp ($^{\circ}$C)& Thermal cond. (Wm$^{-1}$K$^{-1}$) & K\_VRH (GPa) \\
    \midrule
    \multicolumn{4}{l}{\textit{Random Forest evaluation:}} \\
    Site Stats Fingerprint   & $415.7\pm2.2$             & $ 5.41\pm0.02$            & $43.02\pm0.11$ \\
    Sine Coulomb Fingerprint & $336.7\pm2.2$             & $3.50\pm0.03$ & $25.74\pm0.13$ \\
    Element Properties       & $\textbf{106.1}\pm1.6$    & $\textbf{3.39}\pm0.03$    & $\textbf{17.64}\pm0.07$ \\
    CrystalCLR (ours)         & $  161.7\pm2.5$           & $ 3.73\pm0.03$            & $21.98\pm{0.11}$     \\
    CrystalCLR + CS (ours)    & $  156.7\pm1.8$          & $3.51\pm0.03$             & $20.79\pm0.07$ \\
    \midrule
    \multicolumn{4}{l}{\textit{Fine-tune:}} \\
    Random init              & $119.9\pm2.0$             & $3.08\pm0.16$             & $13.13\pm0.25$    \\
    CrystalCLR (ours)         & $106.8\pm6.0$             & $\textbf{3.04}\pm0.33$    & $\textbf{12.89}\pm0.19$    \\
    CrystalCLR + CS (ours)    & $\textbf{102.8}\pm2.6$    & $3.11\pm0.13$    & $12.95\pm0.28$    \\
    \bottomrule
  \end{tabular}
  \vspace{2mm}
  \caption{Evaluation of models, measured in mean absolute
  error. Each reported value is the mean over five random seeds, along with standard deviation}
  \label{table:results}
  \vspace{-0.8cm}
\end{table}

\section{Conclusion}

In this work, we introduce a method for the contrastive training of graph neural networks for
representation learning of materials. First, we establish several crystalline material specific
transformations, and study the effects of the composition of transformations on representation 
quality. In addition, we introduce a novel loss function to explicitly maximize the similarity among materials that share common elements. We show that our framework is
capable of producing representations competitive with engineered material
fingerprinting techniques. 
Finally we show that constrastive pretraining improves performance over random initialization for
downstream tasks. 
Future work will investigate additional contrastive learning methods, the ability to use CrystalCLR representations for material retrieval, and the applicability of equivariant graph neural networks. 

\begin{ack}
DISTRIBUTION STATEMENT A. Approved for public release. Distribution is unlimited. This material is based upon work supported by the Under Secretary of Defense for Research and Engineering under Air Force Contract No. FA8702-15-D-0001. Any opinions, findings, conclusions or recommendations expressed in this material are those of the author(s) and do not necessarily reflect the views of the Under Secretary of Defense for Research and Engineering.© 2022 Massachusetts Institute of Technology. Delivered to the U.S. Government with Unlimited Rights, as defined in DFARS Part 252.227-7013 or 7014 (Feb 2014). Notwithstanding any copyright notice, U.S. Government rights in this work are defined by DFARS 252.227-7013 or DFARS 252.227-7014 as detailed above. Use of this work other than as specifically authorized by the U.S. Government may violate any copyrights that exist in this work.
\end{ack}

{
\small

\bibliography{references}

\begin{thebibliography}{26}
\providecommand{\natexlab}[1]{#1}
\providecommand{\url}[1]{\texttt{#1}}
\expandafter\ifx\csname urlstyle\endcsname\relax
  \providecommand{\doi}[1]{doi: #1}\else
  \providecommand{\doi}{doi: \begingroup \urlstyle{rm}\Url}\fi

\bibitem[Schleder et~al.(2019)Schleder, Padilha, Acosta, Costa, and
  Fazzio]{Schleder2019}
Gabriel~R. Schleder, Antonio C.~M. Padilha, Carlos~Mera Acosta, Marcio Costa,
  and Adalberto Fazzio.
\newblock From dft to machine learning: recent approaches to materials
  science–a review.
\newblock \emph{Journal of Physics: Materials}, 2\penalty0 (3):\penalty0
  032001, 2019.

\bibitem[Ramprasad et~al.(2017)Ramprasad, Batra, Pilania, Mannodi-Kanakkithodi,
  and Kim]{Ramprasad2017}
Rampi Ramprasad, Rohit Batra, Ghanshyam Pilania, Arun Mannodi-Kanakkithodi, and
  Chiho Kim.
\newblock Machine learning in materials informatics: recent applications and
  prospects.
\newblock \emph{npj Computational Materials}, 3\penalty0 (54), 2017.

\bibitem[Xie and Grossman(2018)]{xie2018crystal}
Tian Xie and Jeffrey~C Grossman.
\newblock Crystal graph convolutional neural networks for an accurate and
  interpretable prediction of material properties.
\newblock \emph{Physical review letters}, 120\penalty0 (14):\penalty0 145301,
  2018.

\bibitem[Chen et~al.(2019)Chen, Ye, Zuo, Zheng, and Ong]{Chen2019}
Chi Chen, Weike Ye, Yunxing Zuo, Chen Zheng, and Shyue~Ping Ong.
\newblock Graph networks as a universal machine learning framework for
  molecules and crystals.
\newblock \emph{Chemistry of Materials}, 31\penalty0 (9):\penalty0 3564--3572,
  2019.
\newblock \doi{10.1021/acs.chemmater.9b01294}.
\newblock URL \url{https://doi.org/10.1021/acs.chemmater.9b01294}.

\bibitem[Louis et~al.(2020)Louis, Zhao, Nasiri, Wang, Song, Liu, and
  Hu]{Louis2020}
Steph-Yves Louis, Yong Zhao, Alireza Nasiri, Xiran Wang, Yuqi Song, Fei Liu,
  and Jianjun Hu.
\newblock Graph convolutional neural networks with global attention for
  improved materials property prediction.
\newblock \emph{Physical Chemistry Chemical Physics}, 22\penalty0
  (32):\penalty0 18141–18148, 2020.
\newblock ISSN 1463-9084.
\newblock \doi{10.1039/d0cp01474e}.
\newblock URL \url{http://dx.doi.org/10.1039/D0CP01474E}.

\bibitem[Griggs et~al.(2020)Griggs, Li, and Caceres]{griggs2020unified}
Petar Griggs, Lin Li, and Rajmonda Caceres.
\newblock Unified gnn architecture design for high-throughput material
  screening.
\newblock 2020.

\bibitem[Batzner et~al.(2022)Batzner, Musaelian, Sun, Geiger, Mailoa,
  Kornbluth, Molinari, Smidt, and Kozinsky]{batzner20223}
Simon Batzner, Albert Musaelian, Lixin Sun, Mario Geiger, Jonathan~P Mailoa,
  Mordechai Kornbluth, Nicola Molinari, Tess~E Smidt, and Boris Kozinsky.
\newblock E (3)-equivariant graph neural networks for data-efficient and
  accurate interatomic potentials.
\newblock \emph{Nature communications}, 13\penalty0 (1):\penalty0 1--11, 2022.

\bibitem[Musaelian et~al.(2022)Musaelian, Batzner, Johansson, Sun, Owen,
  Kornbluth, and Kozinsky]{musaelian2022learning}
Albert Musaelian, Simon Batzner, Anders Johansson, Lixin Sun, Cameron~J Owen,
  Mordechai Kornbluth, and Boris Kozinsky.
\newblock Learning local equivariant representations for large-scale atomistic
  dynamics.
\newblock \emph{arXiv preprint arXiv:2204.05249}, 2022.

\bibitem[Magar et~al.(2021)Magar, Wang, Lorsung, Liang, Ramasubramanian, Li,
  and Farimani]{magar2021auglichem}
Rishikesh Magar, Yuyang Wang, Cooper Lorsung, Chen Liang, Hariharan
  Ramasubramanian, Peiyuan Li, and Amir~Barati Farimani.
\newblock Auglichem: Data augmentation library ofchemical structures for
  machine learning, 2021.

\bibitem[Fung et~al.(2021)Fung, Zhang, Juarez, and
  Sumpter]{fung2021benchmarking}
Victor Fung, Jiaxin Zhang, Eric Juarez, and Bobby~G Sumpter.
\newblock Benchmarking graph neural networks for materials chemistry.
\newblock \emph{npj Computational Materials}, 7\penalty0 (1):\penalty0 1--8,
  2021.

\bibitem[Lee and Asahi(2021)]{lee2021transfer}
Joohwi Lee and Ryoji Asahi.
\newblock Transfer learning for materials informatics using crystal graph
  convolutional neural network.
\newblock \emph{Computational Materials Science}, 190:\penalty0 110314, 2021.

\bibitem[Chen et~al.(2020)Chen, Kornblith, Norouzi, and Hinton]{chen2020simple}
Ting Chen, Simon Kornblith, Mohammad Norouzi, and Geoffrey Hinton.
\newblock A simple framework for contrastive learning of visual
  representations.
\newblock In \emph{International conference on machine learning}, pages
  1597--1607. PMLR, 2020.

\bibitem[You et~al.(2020)You, Chen, Sui, Chen, Wang, and Shen]{you2020graph}
Yuning You, Tianlong Chen, Yongduo Sui, Ting Chen, Zhangyang Wang, and Yang
  Shen.
\newblock Graph contrastive learning with augmentations.
\newblock \emph{Advances in Neural Information Processing Systems},
  33:\penalty0 5812--5823, 2020.

\bibitem[Wang et~al.(2022)Wang, Wang, Cao, and
  Barati~Farimani]{wang2022molecular}
Yuyang Wang, Jianren Wang, Zhonglin Cao, and Amir Barati~Farimani.
\newblock Molecular contrastive learning of representations via graph neural
  networks.
\newblock \emph{Nature Machine Intelligence}, 4\penalty0 (3):\penalty0
  279--287, 2022.

\bibitem[Khosla et~al.(2020)Khosla, Teterwak, Wang, Sarna, Tian, Isola,
  Maschinot, Liu, and Krishnan]{khosla2020supervised}
Prannay Khosla, Piotr Teterwak, Chen Wang, Aaron Sarna, Yonglong Tian, Phillip
  Isola, Aaron Maschinot, Ce~Liu, and Dilip Krishnan.
\newblock Supervised contrastive learning.
\newblock \emph{Advances in Neural Information Processing Systems},
  33:\penalty0 18661--18673, 2020.

\bibitem[Sohn(2016)]{sohn2016improved}
Kihyuk Sohn.
\newblock Improved deep metric learning with multi-class n-pair loss objective.
\newblock \emph{Advances in neural information processing systems}, 29, 2016.

\bibitem[Jain et~al.(2013)Jain, Ong, Hautier, Chen, Richards, Dacek, Cholia,
  Gunter, Skinner, Ceder, et~al.]{jain2013commentary}
Anubhav Jain, Shyue~Ping Ong, Geoffroy Hautier, Wei Chen, William~Davidson
  Richards, Stephen Dacek, Shreyas Cholia, Dan Gunter, David Skinner, Gerbrand
  Ceder, et~al.
\newblock Commentary: The materials project: A materials genome approach to
  accelerating materials innovation.
\newblock \emph{APL materials}, 1\penalty0 (1):\penalty0 011002, 2013.

\bibitem[Gao et~al.(2013)Gao, Liu, et~al.]{gao2013recent}
Zhi-Yu Gao, Guo-Quan Liu, et~al.
\newblock Recent progress of web-enable material database and a case study of
  nims and matweb.
\newblock \emph{Journal of Materials Engineering}, 3\penalty0 (11):\penalty0
  89--96, 2013.

\bibitem[Curtarolo et~al.(2012)Curtarolo, Setyawan, Wang, Xue, Yang, Taylor,
  Nelson, Hart, Sanvito, Buongiorno-Nardelli, Mingo, and Levy]{Curtarolo2012}
Stefano Curtarolo, Wahyu Setyawan, Shidong Wang, Junkai Xue, Kesong Yang,
  Richard~H. Taylor, Lance~J. Nelson, Gus~L.W. Hart, Stefano Sanvito, Marco
  Buongiorno-Nardelli, Natalio Mingo, and Ohad Levy.
\newblock Aflowlib.org: A distributed materials properties repository from
  high-throughput ab initio calculations.
\newblock \emph{Computational Materials Science}, 58:\penalty0 227--235, 2012.
\newblock ISSN 0927-0256.
\newblock \doi{https://doi.org/10.1016/j.commatsci.2012.02.002}.
\newblock URL
  \url{https://www.sciencedirect.com/science/article/pii/S0927025612000687}.

\bibitem[Kingma and Ba(2014)]{kingma2014adam}
Diederik~P Kingma and Jimmy Ba.
\newblock Adam: A method for stochastic optimization.
\newblock \emph{arXiv preprint arXiv:1412.6980}, 2014.

\bibitem[Breiman(2001)]{breiman2001random}
Leo Breiman.
\newblock Random forests.
\newblock \emph{Machine learning}, 45\penalty0 (1):\penalty0 5--32, 2001.

\bibitem[Zimmermann and Jain(2020)]{zimmermann2020local}
Nils~ER Zimmermann and Anubhav Jain.
\newblock Local structure order parameters and site fingerprints for
  quantification of coordination environment and crystal structure similarity.
\newblock \emph{RSC advances}, 10\penalty0 (10):\penalty0 6063--6081, 2020.

\bibitem[Faber et~al.(2015)Faber, Lindmaa, von Lilienfeld, and
  Armiento]{faber2015crystal}
Felix Faber, Alexander Lindmaa, O~Anatole von Lilienfeld, and Rickard Armiento.
\newblock Crystal structure representations for machine learning models of
  formation energies.
\newblock \emph{International Journal of Quantum Chemistry}, 115\penalty0
  (16):\penalty0 1094--1101, 2015.

\bibitem[Ward et~al.(2016)Ward, Agrawal, Choudhary, and
  Wolverton]{ward2016general}
Logan Ward, Ankit Agrawal, Alok Choudhary, and Christopher Wolverton.
\newblock A general-purpose machine learning framework for predicting
  properties of inorganic materials.
\newblock \emph{npj Computational Materials}, 2\penalty0 (1):\penalty0 1--7,
  2016.

\bibitem[Pedregosa et~al.(2011)Pedregosa, Varoquaux, Gramfort, Michel, Thirion,
  Grisel, Blondel, Prettenhofer, Weiss, Dubourg, Vanderplas, Passos,
  Cournapeau, Brucher, Perrot, and Duchesnay]{scikit-learn}
F.~Pedregosa, G.~Varoquaux, A.~Gramfort, V.~Michel, B.~Thirion, O.~Grisel,
  M.~Blondel, P.~Prettenhofer, R.~Weiss, V.~Dubourg, J.~Vanderplas, A.~Passos,
  D.~Cournapeau, M.~Brucher, M.~Perrot, and E.~Duchesnay.
\newblock Scikit-learn: Machine learning in {P}ython.
\newblock \emph{Journal of Machine Learning Research}, 12:\penalty0 2825--2830,
  2011.

\bibitem[McInnes et~al.(2018)McInnes, Healy, and Melville]{mcinnes2018umap}
Leland McInnes, John Healy, and James Melville.
\newblock Umap: Uniform manifold approximation and projection for dimension
  reduction.
\newblock \emph{arXiv preprint arXiv:1802.03426}, 2018.

\end{thebibliography}

}

\newpage
\section{Appendix}

\subsection{Element Properties List}

Table \ref{table:element_properties} lists the element properties used for the Element Property classifier.

%
%

\begin{table}[h]
  \centering
  \begin{tabular}{p{0.25\linewidth}p{0.65\linewidth}}
    \toprule
        Feature Name & Description \\
    \midrule
        \texttt{Number} & Atomic Number\\
        \texttt{MendeleevNumber} & Mendeleev Number (position on the periodic table, counting columnwise from H)\\
        \texttt{AtomicWeight} & Atomic weight\\
        \texttt{MeltingT} & Melting temperature of element\\
        \texttt{Column} & Column on periodic table \\
        \texttt{Row} & Row on periodic table \\
        \texttt{CovalentRadius} & Covalent radius of each element\\
        \texttt{Electronegativity} & Pauling electronegativity \\
        \texttt{NsValence} & Number of filled s valence orbitals\\
        \texttt{NpValence} & Number of filled p valence orbitals\\
        \texttt{NdValence} & Number of filled d valence orbitals\\
        \texttt{NfValence} & Number of filled f valence orbitals\\
        \texttt{NValence} & Number of valence electrons\\
        \texttt{NsUnfilled} & Number of unfilled s valence orbitals\\
        \texttt{NpUnfilled} & Number of unfilled p valence orbitals \\
        \texttt{NdUnfilled} & Number of unfilled d valence orbitals\\
        \texttt{NfUnfilled} & Number of unfilled f valence orbitals \\
        \texttt{NUnfilled} & Number of unfilled valence orbitals\\
        \texttt{GSvolume\_pa} & DFT volume per atom of T=0K ground state\\
        \texttt{GSbandgap} & DFT bandgap energy of T=0K ground state\\
        \texttt{GSmagmom} & DFT magnetic moment of T=0K ground state\\
        \texttt{SpaceGroupNumber} & Space group of T=0K ground state structure \\
    \bottomrule
  \end{tabular}
  \vspace{2mm}
  \caption{Element properties used. Feature names and descriptions are reproduced from the supplementary materials of \citet{ward2016general}.}
  \label{table:element_properties}
\end{table}

\end{document}